\documentclass[11pt]{article}

\usepackage{acl}
\usepackage{times}
\usepackage{latexsym}

\usepackage[T1]{fontenc}

\usepackage[utf8]{inputenc}

\usepackage{multirow}

\usepackage{microtype}

\usepackage{inconsolata}

\usepackage{graphicx}
\usepackage{amsmath}
\usepackage{booktabs}
\graphicspath{{figures/}{../figures/}}

\title{Penalty-Framed No-Valid-Option MCQA: Analyzing LLM Abstention under Invalid Choices}

\author{Jinhyeok Kim \\
  Applied Artificial Intelligence \\
  Hanyang University, Republic of Korea \\
  \texttt{jharmyend929@hanyang.ac.kr} \\\And
  Hye-Young Jung\textsuperscript{$\dagger$} \\
  Mathematical Data Science \\
  Hanyang University, Republic of Korea \\
  \texttt{hyjunglove@hanyang.ac.kr} \\
}

\newcommand{\std}[1]{\,\scalebox{0.75}{\ensuremath{\pm\,#1}}}

\begin{document}
\maketitle

\begingroup
\renewcommand{\thefootnote}{\fnsymbol{footnote}}
\footnotetext[2]{Corresponding author}
\endgroup

\begin{abstract}
Multiple-choice question answering (MCQA) is commonly used to evaluate large language models under the assumption that one of the provided options is correct, typically using answer-selection accuracy. However, in real deployments, users or retrieval systems may provide invalid option sets in which none of the listed choices is correct, and selecting one of them may incur downstream cost. We study this setting as penalty-framed no-valid-option MCQA. Using the mathematics subset of MMLU-Pro, we remove the labeled correct option, allow models to either choose a remaining option or output \textsc{Abstain}, and penalize invalid forced-choice responses. We further introduce correct-conditioned analysis, evaluating abstention only on instances that the model originally answered correctly. Experiments show that high MCQA accuracy does not fully guarantee abstention reliability: even under explicit no-valid-option-aware instructions and penalty-based scoring, models still produce invalid forced-choice responses for a subset of originally correct instances. These results show that penalty-framed no-valid-option MCQA reveals an aspect of model reliability not captured by standard answer-selection accuracy.
\end{abstract}

\section{Introduction}

Multiple-choice question answering (MCQA) is widely used to evaluate large language models (LLMs), as it provides a fixed answer space and enables simple automatic scoring \citep{hendrycks2021mmlu,wang2024mmlupro,li2024can}. In the standard MCQA setting, a model receives a question and a set of candidate options, and its performance is measured by whether it selects the labeled correct answer. This protocol has been useful for comparing the knowledge and reasoning abilities of LLMs, but it is usually designed for settings in which a valid answer is present among the provided options.

This design choice matters because answer selection and option-set rejection are not the same capability \citep{goral-etal-2025-wait}. A model may correctly select the answer when it is listed, yet behave differently when the valid option is removed. This motivates a targeted evaluation question: does standard MCQA success transfer to reliable abstention when no valid option remains?

Recent work has studied related problems from two directions. First, work on no-valid-option MCQA asks whether models can recognize that none of the provided options is correct. \textit{Wait, that's not an option} frames this ability as reflective judgment, showing that many aligned models still select invalid options when all listed choices are wrong \citep{goral-etal-2025-wait}. Related work on MCQA reliability similarly suggests that LLMs may select the “least incorrect” option rather than treat the selected answer as uniquely correct \citep{wang-etal-2025-llms-may}, raising concerns about what standard MCQA accuracy measures \citep{zheng2024large,balepur2024artifacts}.

Second, work on abstention under penalty-based scoring studies whether LLMs adjust answer-versus-abstain decisions when wrong answers incur penalties \citep{wen2025know}. RiskEval evaluates whether models adapt their answer-versus-abstain decisions as the penalty for incorrect answers changes, and finds that models often fail to translate uncertainty or verbal confidence into strategic abstention, even when high penalties make abstention preferable \citep{wang2026faithful}.

These two directions motivate the setting we study: what should an LLM do when no provided option is valid and selecting any remaining option is penalized? Rather than treating no-valid-option recognition and abstention under penalty-based scoring as separate problems, we combine them into a targeted MCQA stress test. In this setting, the model must combine two abilities: recognizing that the option set contains no valid answer and avoiding a costly forced-choice response. We call this setting penalty-framed no-valid-option MCQA.

In our setup, we start from standard MCQA instances and remove the labeled correct option. The model is then allowed to either select one of the remaining options or output \textsc{Abstain}. Since the remaining options are intended to be invalid, selecting any of them incurs a penalty, while abstention receives zero utility. Under this scoring rule, abstention is the utility-maximizing action. This setting evaluates whether LLMs can reject invalid option sets when selecting a listed option is penalized, rather than merely selecting the best available option.

However, overall abstention rate alone is insufficient for interpreting model behavior. If a model selects a remaining option after correct-option removal, it may be because the model did not know the original answer, not necessarily because it failed to recognize that the valid option disappeared. To address this ambiguity, we introduce correct-conditioned analysis. We first identify instances that the model answers correctly in the original MCQA setting, and then evaluate the same instances after removing the correct option. This allows us to ask whether success in standard MCQA transfers to reliable abstention under penalty-framed no-valid-option conditions.

Our contributions are as follows. First, we define penalty-framed no-valid-option MCQA, an evaluation setting where no provided option is valid and selecting an invalid option incurs penalty. Second, we introduce correct-conditioned analysis as a way to measure accuracy-to-rejection transfer: whether a model’s demonstrated success on an originally answerable MCQA instance transfers to rejecting the corresponding post-exclusion option set. This separates lack of original task knowledge from failure to reject invalid options after demonstrated answer-selection success. Third, we show that high MCQA accuracy does not fully guarantee abstention reliability: even among originally correct instances, LLMs can still produce invalid forced-choice responses under penalty-based scoring.

\section{Evaluation Setting}
\paragraph{Penalty-Framed No-Valid-Option MCQA.}

A standard MCQA instance consists of a question stem, candidate options, and a labeled correct option. In the standard MCQA evaluation setting, the model is required to select the correct option from the given option set. This implicitly assumes an answerability assumption: at least one valid option is present among the listed choices.

Penalty-framed no-valid-option MCQA targets cases where this answerability condition no longer holds. Given an answerable MCQA instance, we remove the labeled correct option while keeping the question stem and remaining options unchanged. The resulting option set is intended to contain no valid answer. Since equivalent or ambiguous remaining options may violate this condition, our evaluation uses a filtered post-exclusion subset that better satisfies the no-valid-option condition. The model must then either select one of the remaining options or output \textsc{Abstain}.

We evaluate this setting with a penalty-based scoring rule \citep{chow1970optimum,charoenphakdee2021classification}. In standard MCQA, performance is evaluated by whether the model selects the correct option, and selecting an incorrect option is simply treated as an incorrect response. In contrast, a penalty-based scoring rule generally assigns positive score to selecting a correct option, negative penalty to selecting an incorrect option, and zero score to \textsc{Abstain}. However, in the filtered post-exclusion subset, the labeled correct option has been removed, and the remaining selectable options are treated as invalid for evaluation. Selecting any remaining option is therefore treated as an invalid forced-choice response and incurs the negative penalty, whereas \textsc{Abstain} receives zero score. Thus, under this scoring rule, the desired behavior is not to force a choice among the most plausible remaining options, but to reject the option set itself and output \textsc{Abstain}.

\paragraph{Correct-conditioned analysis.}
Overall abstention behavior on no-valid-option instances can conflate cases where the model lacks the original answer with cases where it does not reject the modified option set. To reduce this ambiguity, we perform correct-conditioned analysis: we first identify instances that the model answers correctly in the original MCQA setting, and then evaluate abstention behavior on the corresponding no-valid-option instances.

In the main analysis, we use a stricter stable-correct subset. For each model, an instance is included only if the model answers it correctly in all repeated original MCQA runs. This reduces the chance of including instances that were guessed correctly in a single run and focuses the analysis on whether stable answer-selection success transfers to reliable option-set rejection.

\paragraph{Metrics.}

We use five metrics to evaluate original MCQA performance and behavior under the penalty-framed no-valid-option setting. 
\textbf{Acc} measures answer-selection accuracy over completed responses
in the original MCQA condition, before correct-option removal.

For no-valid-option instances, we report \textbf{AR} and \textbf{IFR}. 
\textbf{AR}, or abstention rate, is the percentage of evaluation-eligible responses in which the model outputs \textsc{Abstain} after the correct option is removed. 
\textbf{IFR}, or invalid forced-choice rate, is the percentage of evaluation-eligible responses in which the model selects one of the remaining options. 
Since the constructed option set contains no valid answer, any such selection is treated as a penalty-incurring forced-choice response. 
Under our parsed response space, each eligible no-valid-option response is classified as either \textsc{Abstain} or an invalid option selection, so AR + IFR = 100\%.

For correct-conditioned analysis, we report \textbf{CAR} and \textbf{CIFR} on each model’s stable-correct subset. \textbf{CAR} measures abstention on instances the model solved consistently before correct-option removal, while \textbf{CIFR} measures invalid forced-choice behavior on the same instances. Thus, AR/IFR characterize overall no-valid-option behavior, whereas CAR/CIFR evaluate whether answer-selection success in standard MCQA transfers to reliable option-set rejection.

\section{Experimental Setup}
\paragraph{Dataset.}

For our main experiments, we use the 10-option mathematics subset of MMLU-Pro \citep{wang2024mmlupro} as a controlled testbed because mathematical answers are relatively verifiable. Since equivalent answer forms or ambiguous options may remain after removing the labeled correct option, we report main results on a filtered post-exclusion subset of 1,118
instances to reduce cases where an alternative valid answer or ambiguous remaining option would be counted as an invalid forced-choice response. Filtering details and examples are provided in Appendix~\ref{app:filtering}.

As a supplementary evaluation beyond mathematics, we also use
the 4-option 400-item multi-domain MMLU \citep{hendrycks2021mmlu} subset from
\citet{goral-etal-2025-wait}.
Applying the same conservative filtering criterion yields
340 instances.

\paragraph{Models.}
Our core experiments on MMLU-Pro Mathematics use the six models
reported in Table~\ref{tab:ablations}: GPT-5-mini, GPT-5-nano,
GPT-4.1-mini, GPT-4.1-nano, Gemini-3.0-Flash, and
Gemini-3.1-Flash-Lite. Under the main setting only, we additionally evaluate GPT-5, Claude Sonnet 5, and the open-weight Qwen3-8B~\citep{yang2025qwen3} on the same mathematics subset. We further evaluate GPT-5-mini and
Qwen3-8B on the multi-domain MMLU subset as a supplementary
evaluation beyond mathematics. Gemini 2.5-family diagnostics
are reported in Appendix~\ref{app:additional-results} because
their outputs exhibited weaker parsing and completion reliability
under our current evaluation pipeline.

\paragraph{Prompts, Scoring Rules, and Runs.}

In the main experiments, we use chain-of-thought prompting \citep{wei2022cot}; direct prompting results are reported in Appendix~\ref{app:direct-prompting-results}. 
The main no-valid-option setting uses a $+4/-1/0$ scoring rule for correct answers, invalid forced-choice responses, and \textsc{Abstain}, respectively, and uses the no-valid-option-aware instruction, which explicitly tells the model to abstain when no listed option is valid.

For ablations, we vary the invalid forced-choice penalty among $\{-0.5,-1,-2\}$ while keeping the no-valid-option-aware instruction fixed, and compare three abstention instructions while keeping the scoring rule fixed: \textit{penalty-only}, \textit{uncertainty-aware} \citep{lin2022teaching,yona2024can}, and \textit{no-valid-option-aware}. 
We use robust parsing and report 3-run means, motivated by concerns about stochastic evaluation reliability \citep{lior2025reliableeval}. 
Full prompts, scoring profiles, run protocol, and parsing details are provided in Appendices~\ref{app:model-settings}, \ref{app:prompts}, and \ref{app:parsing}.
\begin{figure}[t]
  \centering
  \includegraphics[width=\linewidth]{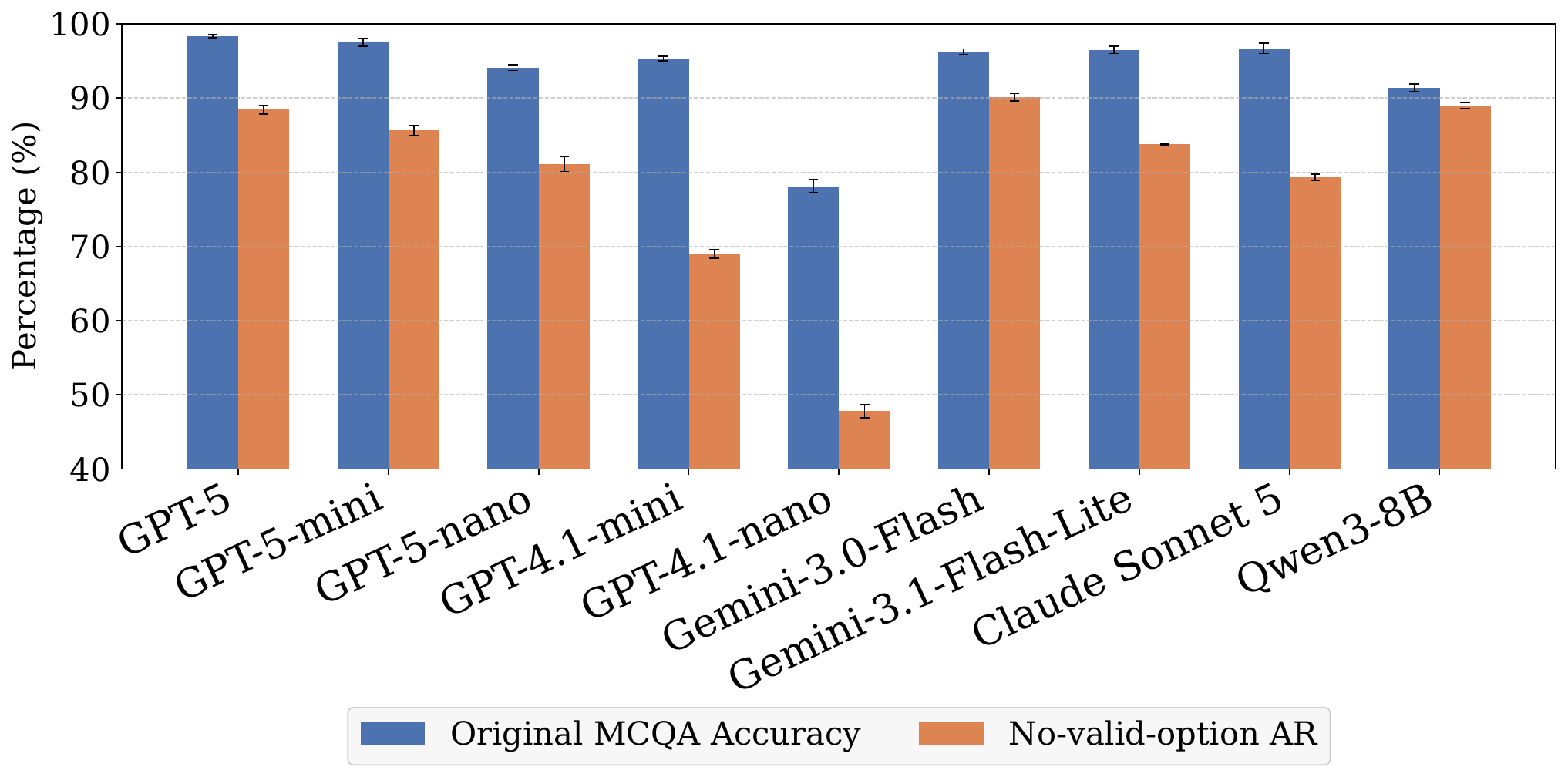}
  \caption{
  Original MCQA accuracy and no-valid-option abstention rate. AR is reported under the main setting: CoT prompting, no-valid-option-aware instruction, invalid forced-choice penalty $-1$, robust parsing. Values are 3-run means on the filtered MMLU-Pro Mathematics subset (1,118 instances).
  }
  \label{fig:accuracy-abstention}
\end{figure}

\section{Results}

\paragraph{High MCQA Accuracy Does Not Fully Guarantee Abstention Reliability.}

Figure~\ref{fig:accuracy-abstention} compares original MCQA accuracy with no-valid-option abstention rate under the main setting. Although the evaluated models often achieve high accuracy on the original answerable MCQA instances, their abstention rates after correct-option removal do not simply mirror original accuracy, with Acc–AR gaps ranging from 2.4 to 30.3 percentage points. This suggests that selecting the correct answer when it is present and abstaining when the valid answer is absent are related but distinct behaviors.

\begin{table}[!ht]
\centering
\small
\resizebox{\columnwidth}{!}{
\begin{tabular}{llccccc}
\toprule
Dataset & Model & Acc$\uparrow$ & AR$\uparrow$
& IFR$\downarrow$ & CAR$\uparrow$ & CIFR$\downarrow$ \\
\midrule
\multirow{9}{*}{\shortstack[l]{MMLU-Pro\\Math}}
& GPT-5-mini            & 97.5 & 85.6 & 14.4 & 87.2 & 12.8 \\
& GPT-5-nano            & 94.1 & 81.1 & 18.9 & 84.5 & 15.5 \\
& GPT-4.1-mini          & 95.3 & 69.0 & 31.0 & 72.4 & 27.6 \\
& GPT-4.1-nano          & 78.1 & 47.8 & 52.2 & 54.5 & 45.5 \\
& Gemini-3.0-Flash      & 96.2 & 90.1 &  9.9 & 90.9 &  9.1 \\
& Gemini-3.1-Flash-Lite & 96.5 & 83.8 & 16.2 & 86.4 & 13.6 \\
\cmidrule(lr){2-7}
& GPT-5                 & 98.3 & 88.4 & 11.6 & 89.8 & 10.2 \\
& Claude Sonnet 5       & 96.7 & 79.3 & 20.7 & 80.6 & 19.4 \\
& Qwen3-8B              & 91.4 & 89.0 & 11.0 & 91.4 &  8.6 \\
\midrule
\multirow{2}{*}{\shortstack[l]{Multi-domain\\MMLU}}
& GPT-5-mini            & 94.0 & 53.5 & 46.5 & 57.8 & 42.2 \\
& Qwen3-8B              & 75.9 & 59.7 & 40.3 & 67.3 & 32.7 \\
\bottomrule
\end{tabular}
}
\caption{
Main post-exclusion results under CoT prompting, robust parsing,
and the no-valid-option-aware instruction with invalid forced-choice 
penalty $-1$.
AR/IFR are computed over all no-valid-option
instances, while CAR/CIFR are computed on each model’s stable-correct subset for the corresponding dataset.
Subset sizes are reported in
Appendix~\ref{app:stable-sizes},
Table~\ref{tab:stable-correct-sizes}.
All values are 3-run means reported as percentages;
standard deviations are reported in
Appendix~\ref{app:filtering},
Table~\ref{tab:unfiltered-filtered}.
}
\label{tab:main-results}
\end{table}

Table~\ref{tab:main-results} further shows that this gap corresponds to non-trivial invalid forced-choice behavior. Even under the no-valid-option-aware instruction, all models have non-zero IFR, ranging from 9.9\% to 52.2\%. Thus, high original MCQA accuracy does not fully guarantee that a model will reject the option set when the correct option is removed and invalid choices are costly.

\paragraph{Invalid Forced Choice Remains on Stably Solved Instances.}
Because stable-correct subsets are model-specific, CAR and CIFR should be interpreted as within-model transfer measures rather than direct comparisons over an identical instance set. Table~\ref{tab:main-results} shows that invalid forced-choice behavior remains even under this stricter condition: CIFR is non-zero even for high-accuracy models such as Gemini-3.0-Flash and GPT-5-mini (9.1\% and 12.8\%), and reaches 45.5\% for GPT-4.1-nano. Among the supplementary models, CIFR is 10.2\% for GPT-5, 19.4\% for Claude Sonnet 5, and 8.6\% for Qwen3-8B. Qwen3-8B also has a substantially lower evaluation-eligible rate in the MMLU-Pro Mathematics no-valid-option condition (76.9\% on average), so its rate-based results should be
interpreted together with the response-quality diagnostics
in Appendix~\ref{sec:response-quality}.

Even when models have demonstrated stable answer-selection success, removing the correct option can still lead them to select an invalid remaining option rather than abstain, revealing a reliability gap between solving an MCQA instance and rejecting its option set.

\paragraph{Supplementary Evidence Beyond Mathematics.}
On the filtered multi-domain MMLU subset, GPT-5-mini and Qwen3-8B show substantially higher invalid forced-choice rates than on MMLU-Pro Mathematics, with IFR of 46.5\% and 40.3\% and stable-correct CIFR of 42.2\% and 32.7\%, respectively. We do not interpret this as a controlled cross-domain comparison, since the two datasets differ in domain composition, size, and the number of remaining options, and only two models were evaluated on the multi-domain subset. Nevertheless, these results provide supplementary evidence that the gap between answer selection and option-set rejection can also appear outside the mathematics setting.

\paragraph{Prompt and Scoring Changes Do Not Eliminate Invalid Forced Choice.}

Table~\ref{tab:ablations} examines whether invalid forced-choice behavior disappears under different scoring rules or abstention instructions. 

\begin{table*}[t]
\centering
\scriptsize
\setlength{\tabcolsep}{2.5pt}
\begin{tabular*}{\textwidth}{@{\extracolsep{\fill}}lcccccc}
\toprule
& \multicolumn{3}{c}{Scoring-rule variation}
& \multicolumn{3}{c}{Abstention-instruction variation} \\
\cmidrule(lr){2-4} \cmidrule(lr){5-7}
Model
& $-0.5$
& $-1$
& $-2$
& Penalty-only
& Uncertainty-aware
& No-valid-option-aware \\
\midrule
GPT-5-mini
& 13.1\std{0.4}
& 12.8\std{0.7}
& 13.2\std{0.8}
& 23.5\std{0.3}
& 21.5\std{1.1}
& 12.8\std{0.7} \\

GPT-5-nano
& 15.2\std{0.4}
& 15.5\std{1.0}
& 14.8\std{0.6}
& 23.4\std{0.8}
& 20.8\std{1.2}
& 15.5\std{1.0} \\

GPT-4.1-mini
& 27.8\std{0.5}
& 27.6\std{0.6}
& 26.6\std{0.2}
& 41.1\std{0.8}
& 48.2\std{0.1}
& 27.6\std{0.6} \\

GPT-4.1-nano
& 46.8\std{0.7}
& 45.5\std{0.6}
& 46.4\std{0.5}
& 75.7\std{0.6}
& 65.0\std{0.4}
& 45.5\std{0.6} \\

Gemini-3.0-Flash
& 10.2\std{0.3}
& 9.1\std{0.6}
& 9.9\std{1.1}
& 31.2\std{0.5}
& 32.5\std{1.4}
& 9.1\std{0.6} \\

Gemini-3.1-Flash-Lite
& 13.5\std{1.0}
& 13.6\std{0.3}
& 12.1\std{0.8}
& 21.7\std{1.3}
& 22.8\std{0.6}
& 13.6\std{0.3} \\

\bottomrule
\end{tabular*}
\caption{
Stable-correct CIFR on the filtered MMLU-Pro Mathematics subset
($N=1{,}118$) under scoring-rule and instruction ablations.
In the scoring-rule block, the no-valid-option-aware instruction is fixed and only the invalid forced-choice penalty varies. In the instruction block, $+4$/$-1$/$0$ scoring is fixed and only the abstention instruction varies. CIFR values are reported as percentages in mean $\pm$ standard deviation format over three independent runs. Lower CIFR is better.
}
\label{tab:ablations}
\end{table*}

Across the tested penalty values, stable-correct CIFR remains within a narrow range for most models. This suggests that residual invalid forced-choice behavior is robust to the particular stated penalty value, rather than being an artifact of a single penalty setting.

Instruction style has a larger effect. The no-valid-option-aware instruction consistently reduces CIFR relative to penalty-only or uncertainty-aware instructions, indicating that explicitly stating the no-valid-option possibility helps reduce costly forced-choice behavior, but every model still has non-zero CIFR under the main setting.

\section{Conclusion}

We studied penalty-framed no-valid-option MCQA, where the labeled correct option is removed and selecting any remaining option incurs a penalty. Using the mathematics subset of MMLU-Pro, we show that high original MCQA accuracy does not fully guarantee reliable abstention after correct-option removal: models still produce invalid forced-choice responses at non-zero rates, even under explicit no-valid-option-aware instructions and on stable-correct instances solved consistently before removal. Supplementary evaluations with additional models on MMLU-Pro Mathematics and with GPT-5-mini and Qwen3-8B on a filtered multi-domain MMLU subset show the same qualitative pattern of
non-zero invalid forced-choice behavior, though these results remain limited to the evaluated datasets and configurations.

Prompt and scoring-rule ablations further show that this behavior can be reduced but not eliminated. These findings suggest that MCQA evaluation should assess not only whether models select correct answers when valid options are present, but also whether they can reject invalid option sets when forced choices are explicitly penalized.

\section*{Limitations}

Our study has several limitations. First, our primary evaluation and all ablation and sensitivity analyses focus on the mathematics subset of MMLU-Pro. We choose this domain because mathematical answers are relatively verifiable, making it suitable for constructing and auditing no-valid-option instances. We additionally report a supplementary evaluation on a filtered 340-item multi-domain MMLU subset with GPT-5-mini and Qwen3-8B, but this extension remains limited in both dataset size and model coverage. Therefore, the multi-domain result should be interpreted as supplementary evidence rather than as a controlled cross-domain comparison. No-valid-option behavior may differ in other domains where answers are less sharply defined, options are paraphrastic, or multiple answers can be partially valid.

Second, model coverage remains limited. The core comparison and all ablations use six closed-source models from two providers; GPT-5, Claude Sonnet 5, and the open-weight Qwen3-8B are evaluated only under the main setting. Our findings should therefore be read as evidence from the evaluated model families and configurations rather than as a universal claim about LLMs.

Third, although we filter cases where a remaining option may still be considered valid, some residual ambiguity may remain. In addition, correct-conditioned metrics are computed on each model’s own stable-correct subset. Thus, CAR and CIFR are best interpreted as within-model measures of whether answer-selection success transfers to option-set rejection, rather than as direct cross-model comparisons over an identical item set.

\section*{Acknowledgments}
This research was supported by Basic Science Research Program through the National Research Foundation of Korea (NRF) funded by the Ministry of Education (NRF-2022R1F1A1074939).

\bibliography{custom}

@inproceedings{goral-etal-2025-wait,
    title = "Wait, that{'}s not an option: {LLM}s Robustness with Incorrect Multiple-Choice Options",
    author = "G{\'o}ral, Gracjan and
      Wi{\'s}nios, Emilia and
      Sankowski, Piotr and
      Budzianowski, Pawe{\l}",
    editor = "Che, Wanxiang and
      Nabende, Joyce and
      Shutova, Ekaterina and
      Pilehvar, Mohammad Taher",
    booktitle = "Proceedings of the 63rd Annual Meeting of the Association for Computational Linguistics (Volume 1: Long Papers)",
    month = jul,
    year = "2025",
    address = "Vienna, Austria",
    publisher = "Association for Computational Linguistics",
    url = "https://aclanthology.org/2025.acl-long.75/",
    doi = "10.18653/v1/2025.acl-long.75",
    pages = "1495--1515",
    ISBN = "979-8-89176-251-0"
}

@inproceedings{wang-etal-2025-llms-may,
    title = "{LLM}s May Perform {MCQA} by Selecting the Least Incorrect Option",
    author = "Wang, Haochun and
      Zhao, Sendong and
      Qiang, Zewen and
      Xi, Nuwa and
      Qin, Bing and
      Liu, Ting",
    editor = "Rambow, Owen and
      Wanner, Leo and
      Apidianaki, Marianna and
      Al-Khalifa, Hend and
      Eugenio, Barbara Di and
      Schockaert, Steven",
    booktitle = "Proceedings of the 31st International Conference on Computational Linguistics",
    month = jan,
    year = "2025",
    address = "Abu Dhabi, UAE",
    publisher = "Association for Computational Linguistics",
    url = "https://aclanthology.org/2025.coling-main.390/",
    pages = "5852--5862"
}

@inproceedings{wang2024mmlupro,
    title = "{MMLU-Pro}: A More Robust and Challenging Multi-Task Language Understanding Benchmark",
    author = "Wang, Yubo and
      Ma, Xueguang and
      Zhang, Ge and
      Ni, Yuansheng and
      Chandra, Abhranil and
      Guo, Shiguang and
      Ren, Weiming and
      Arulraj, Aaran and
      He, Xuan and
      Jiang, Ziyan and
      Li, Tianle and
      Ku, Max and
      Wang, Kai and
      Zhuang, Alex and
      Fan, Rongqi and
      Yue, Xiang and
      Chen, Wenhu",
    booktitle = "Advances in Neural Information Processing Systems",
    volume = "37",
    year = "2024",
    url = "https://proceedings.neurips.cc/paper_files/paper/2024/hash/ad236edc564f3e3156e1b2feafb99a24-Abstract.html",
    doi = "10.52202/079017-3018"
}

@misc{wang2026faithful,
      title={Are {LLM} Decisions Faithful to Verbal Confidence?}, 
      author={Jiawei Wang and Yanfei Zhou and Siddartha Devic and Deqing Fu},
      year={2026},
      eprint={2601.07767},
      archivePrefix={arXiv},
      primaryClass={cs.LG},
      url={https://arxiv.org/abs/2601.07767}, 
}

@inproceedings{hendrycks2021mmlu,
    title = "Measuring Massive Multitask Language Understanding",
    author = "Hendrycks, Dan and
      Burns, Collin and
      Basart, Steven and
      Zou, Andy and
      Mazeika, Mantas and
      Song, Dawn and
      Steinhardt, Jacob",
    booktitle = "International Conference on Learning Representations",
    year = "2021",
    url = "https://openreview.net/forum?id=d7KBjmI3GmQ"
}

@inproceedings{li2024can,
    title = "Can Multiple-choice Questions Really Be Useful in Detecting the Abilities of {LLM}s?",
    author = "Li, Wangyue and
      Li, Liangzhi and
      Xiang, Tong and
      Liu, Xiao and
      Deng, Wei and
      Garcia, Noa",
    booktitle = "Proceedings of the 2024 Joint International Conference on Computational Linguistics, Language Resources and Evaluation",
    year = "2024",
    address = "Torino, Italia",
    publisher = "ELRA and ICCL",
    url = "https://aclanthology.org/2024.lrec-main.251/",
    pages = "2819--2834"
}

@inproceedings{zheng2024large,
    title = "Large Language Models Are Not Robust Multiple Choice Selectors",
    author = "Zheng, Chujie and
      Zhou, Hao and
      Meng, Fandong and
      Zhou, Jie and
      Huang, Minlie",
    booktitle = "International Conference on Learning Representations",
    year = "2024",
    url = "https://openreview.net/forum?id=shr9PXz7T0"
}

@inproceedings{balepur2024artifacts,
    title = "Artifacts or Abduction: How Do {LLM}s Answer Multiple-Choice Questions Without the Question?",
    author = "Balepur, Nishant and
      Ravichander, Abhilasha and
      Rudinger, Rachel",
    booktitle = "Proceedings of the 62nd Annual Meeting of the Association for Computational Linguistics (Volume 1: Long Papers)",
    year = "2024",
    address = "Bangkok, Thailand",
    publisher = "Association for Computational Linguistics",
    url = "https://aclanthology.org/2024.acl-long.555/",
    doi = "10.18653/v1/2024.acl-long.555",
    pages = "10308--10330"
}

@article{chow1970optimum,
    title = "On Optimum Recognition Error and Reject Tradeoff",
    author = "Chow, C. K.",
    journal = "IEEE Transactions on Information Theory",
    volume = "16",
    number = "1",
    pages = "41--46",
    year = "1970",
    doi = "10.1109/TIT.1970.1054406"
}

@inproceedings{charoenphakdee2021classification,
    title = "Classification with Rejection Based on Cost-sensitive Classification",
    author = "Charoenphakdee, Nontawat and
      Cui, Zhenghang and
      Zhang, Yivan and
      Sugiyama, Masashi",
    booktitle = "Proceedings of the 38th International Conference on Machine Learning",
    series = "Proceedings of Machine Learning Research",
    volume = "139",
    pages = "1507--1517",
    year = "2021",
    publisher = "PMLR",
    url = "https://proceedings.mlr.press/v139/charoenphakdee21a.html"
}

@article{wen2025know,
    title = "Know Your Limits: A Survey of Abstention in Large Language Models",
    author = "Wen, Bingbing and
      Yao, Jihan and
      Feng, Shangbin and
      Xu, Chenjun and
      Tsvetkov, Yulia and
      Howe, Bill and
      Wang, Lucy Lu",
    journal = "Transactions of the Association for Computational Linguistics",
    volume = "13",
    pages = "529--556",
    year = "2025",
    url = "https://aclanthology.org/2025.tacl-1.26/",
    doi = "10.1162/tacl_a_00754"
}

@article{lin2022teaching,
    title = "Teaching Models to Express Their Uncertainty in Words",
    author = "Lin, Stephanie and
      Hilton, Jacob and
      Evans, Owain",
    journal = "Transactions on Machine Learning Research",
    year = "2022",
    url = "https://openreview.net/forum?id=8s8K2UZGTZ"
}

@inproceedings{yona2024can,
    title = "Can Large Language Models Faithfully Express Their Intrinsic Uncertainty in Words?",
    author = "Yona, Gal and
      Aharoni, Roee and
      Geva, Mor",
    booktitle = "Proceedings of the 2024 Conference on Empirical Methods in Natural Language Processing",
    year = "2024",
    address = "Miami, Florida, USA",
    publisher = "Association for Computational Linguistics",
    url = "https://aclanthology.org/2024.emnlp-main.443/",
    doi = "10.18653/v1/2024.emnlp-main.443",
    pages = "7752--7764"
}

@inproceedings{wei2022cot,
    title = "Chain-of-Thought Prompting Elicits Reasoning in Large Language Models",
    author = "Wei, Jason and
      Wang, Xuezhi and
      Schuurmans, Dale and
      Bosma, Maarten and
      Ichter, Brian and
      Xia, Fei and
      Chi, Ed H. and
      Le, Quoc V. and
      Zhou, Denny",
    booktitle = "Advances in Neural Information Processing Systems",
    volume = "35",
    pages = "24824--24837",
    year = "2022",
    url = "https://proceedings.neurips.cc/paper_files/paper/2022/hash/9d5609613524ecf4f15af0f7b31abca4-Abstract-Conference.html"
}

@inproceedings{lior2025reliableeval,
    title = "{R}eliable{E}val: A Recipe for Stochastic {LLM} Evaluation via Method of Moments",
    author = "Lior, Gili and
      Habba, Eliya and
      Levy, Shahar and
      Caciularu, Avi and
      Stanovsky, Gabriel",
    booktitle = "Findings of the Association for Computational Linguistics: EMNLP 2025",
    month = nov,
    year = "2025",
    address = "Suzhou, China",
    publisher = "Association for Computational Linguistics",
    url = "https://aclanthology.org/2025.findings-emnlp.594/",
    doi = "10.18653/v1/2025.findings-emnlp.594",
    pages = "11146--11153"
}

@inproceedings{gu2025olmes,
    title = "{OLMES}: A Standard for Language Model Evaluations",
    author = "Gu, Yuling and
      Tafjord, Oyvind and
      Kuehl, Bailey and
      Haddad, Dany and
      Dodge, Jesse and
      Hajishirzi, Hannaneh",
    booktitle = "Findings of the Association for Computational Linguistics: NAACL 2025",
    month = apr,
    year = "2025",
    address = "Albuquerque, New Mexico",
    publisher = "Association for Computational Linguistics",
    url = "https://aclanthology.org/2025.findings-naacl.282/",
    doi = "10.18653/v1/2025.findings-naacl.282",
    pages = "5020--5048"
}

@inproceedings{molfese2025right,
    title = "Right Answer, Wrong Score: Uncovering the Inconsistencies of {LLM} Evaluation in Multiple-Choice Question Answering",
    author = "Molfese, Francesco Maria and
      Moroni, Luca and
      Gioffr{\'e}, Luca and
      Scir{\`e}, Alessandro and
      Conia, Simone and
      Navigli, Roberto",
    booktitle = "Findings of the Association for Computational Linguistics: ACL 2025",
    month = jul,
    year = "2025",
    address = "Vienna, Austria",
    publisher = "Association for Computational Linguistics",
    url = "https://aclanthology.org/2025.findings-acl.950/",
    doi = "10.18653/v1/2025.findings-acl.950",
    pages = "18477--18494"
}

@inproceedings{gema2025done,
    title = "Are We Done with {MMLU}?",
    author = "Gema, Aryo Pradipta and
      Leang, Joshua Ong Jun and
      Hong, Giwon and
      Devoto, Alessio and
      Mancino, Alberto Carlo Maria and
      Saxena, Rohit and
      He, Xuanli and
      Zhao, Yu and
      Du, Xiaotang and
      Madani, Mohammad Reza Ghasemi and
      Barale, Claire and
      McHardy, Robert and
      Harris, Joshua and
      Kaddour, Jean and
      Van Krieken, Emile and
      Minervini, Pasquale",
    booktitle = "Proceedings of the 2025 Conference of the Nations of the Americas Chapter of the Association for Computational Linguistics: Human Language Technologies (Volume 1: Long Papers)",
    month = apr,
    year = "2025",
    address = "Albuquerque, New Mexico",
    publisher = "Association for Computational Linguistics",
    url = "https://aclanthology.org/2025.naacl-long.262/",
    doi = "10.18653/v1/2025.naacl-long.262",
    pages = "5069--5096"
}

@inproceedings{tam-etal-2025-none,
  title = "None of the Above, Less of the Right Parallel Patterns in Human and {LLM} Performance on Multi-Choice Questions Answering",
  author = "Tam, Zhi Rui and Wu, Cheng-Kuang and Lin, Chieh-Yen and Chen, Yun-Nung",
  booktitle = "Findings of the Association for Computational Linguistics: ACL 2025",
  year = "2025",
  address = "Vienna, Austria",
  pages = "20112--20134",
  publisher = "Association for Computational Linguistics",
  url = "https://aclanthology.org/2025.findings-acl.1031/",
  doi = "10.18653/v1/2025.findings-acl.1031"
}

@inproceedings{madhusudhan-etal-2025-llms,
  title = "Do {LLM}s Know When to {NOT} Answer? Investigating Abstention Abilities of Large Language Models",
  author = "Madhusudhan, Nishanth and Madhusudhan, Sathwik Tejaswi and Yadav, Vikas and Hashemi, Masoud",
  booktitle = "Proceedings of the 31st International Conference on Computational Linguistics",
  year = "2025",
  address = "Abu Dhabi, UAE",
  pages = "9329--9345",
  publisher = "Association for Computational Linguistics",
  url = "https://aclanthology.org/2025.coling-main.627/"
}

@misc{yang2025qwen3,
      title={Qwen3 Technical Report}, 
      author={An Yang and Anfeng Li and Baosong Yang and Beichen Zhang and Binyuan Hui and Bo Zheng and Bowen Yu and Chang Gao and Chengen Huang and Chenxu Lv and Chujie Zheng and Dayiheng Liu and Fan Zhou and Fei Huang and Feng Hu and Hao Ge and Haoran Wei and Huan Lin and Jialong Tang and Jian Yang and Jianhong Tu and Jianwei Zhang and Jianxin Yang and Jiaxi Yang and Jing Zhou and Jingren Zhou and Junyang Lin and Kai Dang and Keqin Bao and Kexin Yang and Le Yu and Lianghao Deng and Mei Li and Mingfeng Xue and Mingze Li and Pei Zhang and Peng Wang and Qin Zhu and Rui Men and Ruize Gao and Shixuan Liu and Shuang Luo and Tianhao Li and Tianyi Tang and Wenbiao Yin and Xingzhang Ren and Xinyu Wang and Xinyu Zhang and Xuancheng Ren and Yang Fan and Yang Su and Yichang Zhang and Yinger Zhang and Yu Wan and Yuqiong Liu and Zekun Wang and Zeyu Cui and Zhenru Zhang and Zhipeng Zhou and Zihan Qiu},
      year={2025},
      eprint={2505.09388},
      archivePrefix={arXiv},
      primaryClass={cs.CL},
      url={https://arxiv.org/abs/2505.09388}, 
}

\appendix

\section{Penalty-Framed No-Valid-Option MCQA Construction}
\label{app:construction}

\begin{figure*}[t]
  \centering
  \includegraphics[width=\textwidth]{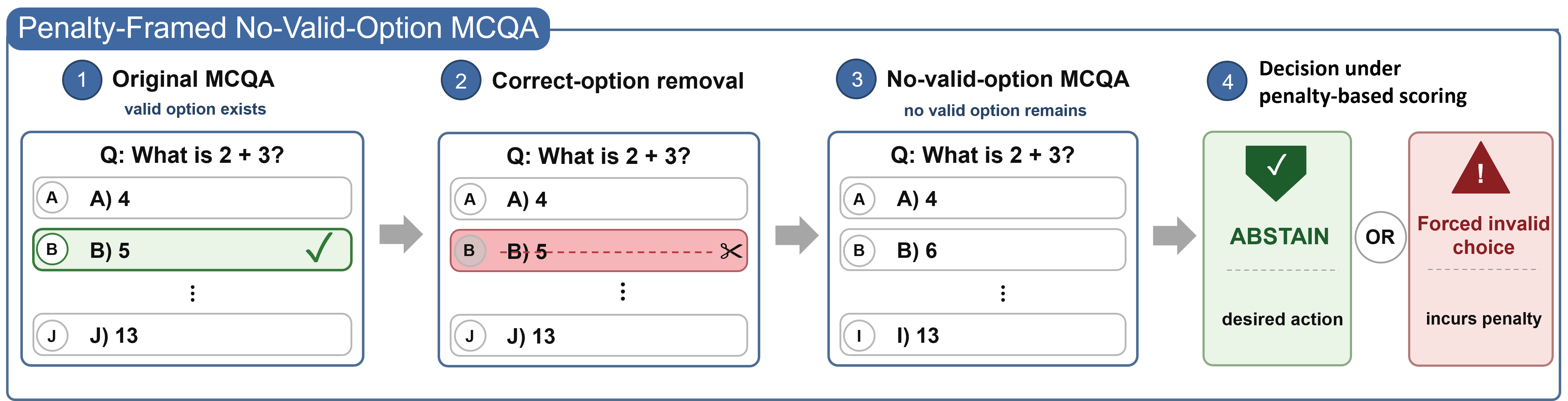}
  \caption{
  Construction of penalty-framed no-valid-option MCQA.
  The labeled correct option is removed from an original MCQA instance.
  The desired behavior is to output \textsc{Abstain}; selecting any remaining option is treated as an invalid forced-choice response and incurs a penalty.
  }
  \label{fig:construction}
\end{figure*}

Figure~\ref{fig:construction} illustrates how we construct no-valid-option instances by removing the labeled correct option from answerable MCQA questions.
After removal, the model must either output \textsc{Abstain} or select one of the remaining options, which is treated as an invalid forced-choice response.

\section{Model and Experimental Settings}
\label{app:model-settings}

This appendix summarizes the models, experimental conditions,
and run protocol used in our experiments.

\subsection{Models}

We evaluate models from OpenAI, Google, Anthropic, and the
open-weight Qwen family. OpenAI models are queried through the
OpenAI Responses API, Gemini models through Google Vertex AI,
Claude Sonnet 5 through the Anthropic API, and Qwen3-8B is run
locally using Hugging Face. Model identifiers and runtime settings are listed in Table~\ref{tab:model-settings}.

Our six core models are GPT-5-mini, GPT-5-nano, GPT-4.1-mini,
GPT-4.1-nano, Gemini-3.0-Flash, and Gemini-3.1-Flash-Lite.
These models are used in the prompting, scoring-rule, and
parser-sensitivity analyses.

Under the main setting, we additionally evaluate GPT-5,
Claude Sonnet 5, and Qwen3-8B on MMLU-Pro Mathematics.
For the supplementary multi-domain MMLU evaluation, we evaluate
GPT-5-mini and Qwen3-8B.

We also evaluate Gemini-2.5-family models as supplementary
diagnostics. Because these models exhibit substantially weaker
completion and parsing reliability under our evaluation pipeline,
we report them separately in Appendix~\ref{app:additional-results}.

\subsection{Experimental Conditions}

We evaluate two task conditions. The first is the original MCQA
condition. In this condition, the labeled correct option remains in
the candidate option set. This condition is used to measure standard
answer-selection accuracy.

The second is the no-valid-option condition. In this condition,
the labeled correct option is removed from the candidate option
set. The model is allowed to either select one of the remaining
options or output \textsc{Abstain}. Since the labeled correct
option has been removed, selecting any remaining option is treated
as an invalid forced-choice response.

For the no-valid-option condition, we evaluate three abstention
instruction styles. The penalty-only instruction specifies the
scoring rule and the expected-score objective, but does not
explicitly state that no listed option may be valid. The
uncertainty-aware instruction tells the model to output
\textsc{Abstain} when it is uncertain. The no-valid-option-aware
instruction explicitly states that none of the listed options may
be valid and instructs the model to output \textsc{Abstain} in
that case.

We also evaluate three penalty-based score profiles. In all
profiles, selecting the correct answer receives $+4$ and
\textsc{Abstain} receives $0$. The penalty for an invalid
forced-choice response is varied among $-0.5$, $-1$, and $-2$.

\subsection{Run Protocol and Token Budgets}

Motivated by recent concerns about the reliability of single-run evaluations for stochastic LLM outputs \citep{lior2025reliableeval}, each model is evaluated with three independent runs. Each dataset instance is queried independently, and no previous model outputs,
conversation state, or thread identifiers are passed between
examples.

We evaluate both direct prompting and chain-of-thought prompting.
The maximum output-token budget is 4096 tokens for direct
prompting and 8192 tokens for chain-of-thought prompting.

For no-valid-option rate-based metrics reported as percentages, we use completed responses with valid parsed predictions. For original MCQA accuracy, completed responses without a
valid parsed option label are counted as incorrect. Incomplete responses
and invalid-format responses are tracked separately and are discussed in
Appendix~\ref{app:parsing}.

\begin{table*}[t]
\centering
\small
\resizebox{\textwidth}{!}{%
\begin{tabular}{llll}
\toprule
Model name in paper & API model identifier & Provider & Runtime setting \\
\midrule

GPT-5-mini
& \texttt{gpt-5-mini-2025-08-07}
& OpenAI
& \texttt{reasoning\_effort=low} \\

GPT-5-nano
& \texttt{gpt-5-nano-2025-08-07}
& OpenAI
& \texttt{reasoning\_effort=low} \\

GPT-4.1-mini
& \texttt{gpt-4.1-mini-2025-04-14}
& OpenAI
& \texttt{temperature=0.0} \\

GPT-4.1-nano
& \texttt{gpt-4.1-nano-2025-04-14}
& OpenAI
& \texttt{temperature=0.0} \\

Gemini-3.0-Flash
& \texttt{gemini-3-flash-preview}
& Google Vertex AI
& \texttt{thinking\_level=low} \\

Gemini-3.1-Flash-Lite
& \texttt{gemini-3.1-flash-lite}
& Google Vertex AI
& \texttt{thinking\_level=low} \\

\midrule

GPT-5
& \texttt{gpt-5-2025-08-07}
& OpenAI
& \texttt{reasoning\_effort=low} \\

Claude Sonnet 5
& \texttt{claude-sonnet-5}
& Anthropic
& \texttt{effort=medium} \\

Qwen3-8B
& \texttt{Qwen/Qwen3-8B}
& Local Hugging Face
& \texttt{temperature=0.6, top\_p=0.95, top\_k=20,} \\
& & &
\texttt{enable\_thinking=True, torch\_dtype=bfloat16} \\

\midrule
Gemini-2.5-Flash
& \texttt{gemini-2.5-flash}
& Google Vertex AI
& \texttt{temperature=0.0}, \texttt{thinking\_budget=0} \\

Gemini-2.5-Flash-Lite
& \texttt{gemini-2.5-flash-lite}
& Google Vertex AI
& \texttt{temperature=0.0}, \texttt{thinking\_budget=0} \\

\bottomrule
\end{tabular}
}
\caption{
Model identifiers and runtime settings used in our experiments.
The six core models are used in the prompting, scoring-rule,
and parser-sensitivity analyses. GPT-5, Claude Sonnet 5, and
Qwen3-8B are additionally evaluated under the main setting.
Gemini 2.5-family models are reported only as supplementary
diagnostics because of weaker completion and parsing reliability
under the current evaluation pipeline.
}
\label{tab:model-settings}
\end{table*}

\section{Filtering Details}
\label{app:filtering}

This appendix describes why post-exclusion filtering is necessary, how we performed it, and how the main results change before and after filtering.

\subsection{Why Filtering Is Necessary}

Our no-valid-option construction removes the labeled correct option
from an original MCQA instance. This construction is valid only when
no remaining option is also a valid answer. However, in mathematical
MCQA, equivalent answer forms may remain after the labeled correct
option is removed. In such cases, treating every remaining option as invalid would risk counting an alternative valid answer as an invalid forced-choice response. This concern is consistent with broader evidence that benchmark labels and answer options can contain nontrivial errors or ambiguities \citep{gema2025done}.

Figure~\ref{fig:filtering-example} shows an example from the
MMLU-Pro mathematics subset \citep{wang2024mmlupro}. The labeled correct option is ``1:04'',
but another remaining option, ``15:60'', can be interpreted as an
equivalent ratio. If the labeled correct option is removed, the
remaining option set may still contain a valid answer. We therefore filter such cases from the main post-exclusion evaluation set.

\begin{figure}[!ht]
\centering
\fbox{
\begin{minipage}{0.92\linewidth}
\small
\textbf{Example instance requiring filtering}

\medskip
\textbf{Question.}
A zoo has 15 toucans and 60 parrots. What is the ratio of the
number of toucans to the number of parrots at the zoo?

\medskip
\textbf{Options.}
1:03, 60:15, 4:01, \textbf{1:04}, 3:01, 15:60, 5:04, 1:05,
4:05, 5:01

\medskip
\textbf{Labeled answer.}
1:04

\medskip
\textbf{Reason for exclusion.}
Although the labeled answer is 1:04, the option 15:60 is equivalent
to the same ratio under a natural interpretation. Therefore, after
removing the labeled answer, the remaining option set may still
contain a valid answer.
\end{minipage}
}
\caption{Example of a case where correct-option removal does not
guarantee a no-valid-option instance.}
\label{fig:filtering-example}
\end{figure}

\subsection{Filtering Procedure}

We used both ChatGPT and Gemini as independent screening assistants
to identify potentially problematic instances. These models were used
only to support candidate identification and explanation. The final
inclusion or exclusion decision was made by an internal audit.

For each batch of instances, we asked the screening models to inspect
the question, options, and labeled answer directly, rather than relying
only on string matching or automatic similarity heuristics. The prompt
used for this screening was as follows.

\begin{quote}
\small
Our goal is to create multiple-choice questions with no correct option
by removing the labeled correct option from existing MCQA instances.

Please directly inspect each problem and answer the following questions.

\begin{enumerate}
    \item Check whether the labeled correct answer in the original
    dataset matches the answer you obtain by solving the problem.
    Report all problems, separating cases where the label appears
    correct from cases where it appears incorrect.

    \item Among the remaining options after excluding the labeled
    correct option, check whether any option is equivalent to the
    labeled correct answer. Report all problems, separating cases with
    equivalent remaining options from cases without them.

    \item After removing the labeled correct option, check whether any
    other remaining option could still be a correct answer. Report all
    problems, separating cases where no remaining option is correct
    from cases where another option could be correct.
\end{enumerate}

Please review the problems one by one. Do not rely only on string
similarity or automatic filtering. For each item, inspect the question,
options, and answer directly.

Please return the output as a table with the following columns:

\medskip
\noindent
\texttt{question\_id | label | decision | reason | recommendation}

\medskip
In the reason column, provide a specific explanation for the decision.

Please make sure to:
\begin{itemize}
    \item review every problem in the batch;
    \item inspect all 20 problems directly, not only automatically
    selected candidates;
    \item summarize the recommended question IDs to exclude;
    \item verify the cases carefully and rigorously.
\end{itemize}
\end{quote}

After this screening process, we manually reviewed the candidate
instances and constructed the final post-exclusion subset. From the
1,237 10-option MMLU-Pro mathematics instances, we selected 1,118
instances for the final filtered post-exclusion subset. The selected
instances are cases where, after removing the labeled correct option,
the remaining option set can be treated as containing no valid answer
with high confidence.

Table~\ref{tab:exclusion-counts} summarizes the exclusion counts
by reason for both datasets.
We excluded 119 of 1,237 MMLU-Pro Mathematics items and
60 of 400 multi-domain MMLU items, retaining 1,118 and 340
items, respectively.

We conservatively excluded items with labeling issues or plausible
alternative correct options that could violate the no-valid-option
premise after removal. Exclusion does not necessarily imply that
the original benchmark item is incorrect or unusable.

\begin{table}[t]
\centering
\small
\begin{tabular}{lrr}
\toprule
Exclusion reason
& \shortstack{MMLU-Pro\\Math}
& \shortstack{Multi-domain\\MMLU} \\
\midrule
Label error                & 47  & 26 \\
Alternative correct option & 72  & 34 \\
\midrule
Total excluded             & 119 & 60 \\
\bottomrule
\end{tabular}
\caption{
Exclusion counts by reason.
Alternative correct options are options that could remain valid
after removal of the labeled correct option.
}
\label{tab:exclusion-counts}
\end{table}

\subsection{Unfiltered vs. Filtered Results}

Table~\ref{tab:unfiltered-filtered} compares the main results before
and after filtering under the same evaluation setting: CoT prompting,
the no-valid-option-aware instruction, invalid forced-choice penalty
$-1$, robust parsing, and three-run means.

Filtering generally increases original MCQA accuracy and abstention
rates while reducing invalid forced-choice rates. This pattern is
consistent with the filtering step reducing the risk that remaining
options contain alternative valid answers or ambiguous answer forms. Importantly, even after this filtering step, all models still exhibit
non-zero CIFR, suggesting that residual invalid forced-choice behavior
is not explained solely by ambiguous unfiltered instances.

\begin{table*}[t]
\centering
\scriptsize
\setlength{\tabcolsep}{1.5pt}
\begin{tabular*}{\textwidth}{@{\extracolsep{\fill}}lllcccccc@{}}
\toprule
Dataset & Model & Set & $N$ & Acc$\uparrow$ & AR$\uparrow$
& IFR$\downarrow$ & CAR$\uparrow$ & CIFR$\downarrow$ \\
\midrule
& GPT-5-mini & Unfiltered & 1237 & 92.3\std{0.6} & 79.8\std{0.6} & 20.2\std{0.6} & 84.3\std{0.6} & 15.7\std{0.6} \\
& & Filtered & 1118 & 97.5\std{0.5} & 85.6\std{0.7} & 14.4\std{0.7} & 87.2\std{0.7} & 12.8\std{0.7} \\
\cmidrule{2-9}
& GPT-5-nano & Unfiltered & 1237 & 89.6\std{0.4} & 76.4\std{0.8} & 23.6\std{0.8} & 81.9\std{0.9} & 18.1\std{0.9} \\
& & Filtered & 1118 & 94.1\std{0.4} & 81.1\std{1.0} & 18.9\std{1.0} & 84.5\std{1.0} & 15.5\std{1.0} \\
\cmidrule{2-9}
& GPT-4.1-mini & Unfiltered & 1237 & 91.3\std{0.2} & 64.3\std{0.6} & 35.7\std{0.6} & 69.3\std{0.5} & 30.7\std{0.5} \\
& & Filtered & 1118 & 95.3\std{0.3} & 69.0\std{0.6} & 31.0\std{0.6} & 72.4\std{0.6} & 27.6\std{0.6} \\
\cmidrule{2-9}
& GPT-4.1-nano & Unfiltered & 1237 & 74.7\std{0.6} & 44.6\std{0.9} & 55.4\std{0.9} & 52.7\std{0.7} & 47.3\std{0.7} \\
& & Filtered & 1118 & 78.1\std{0.9} & 47.8\std{0.9} & 52.2\std{0.9} & 54.5\std{0.6} & 45.5\std{0.6} \\
\cmidrule{2-9}
\multirow{2}{*}{\shortstack[l]{MMLU-Pro\\Math}}
& Gemini-3.0-Flash & Unfiltered & 1237 & 92.7\std{0.3} & 84.0\std{0.7} & 16.0\std{0.7} & 87.0\std{0.6} & 13.0\std{0.6} \\
& & Filtered & 1118 & 96.2\std{0.4} & 90.1\std{0.5} & 9.9\std{0.5} & 90.9\std{0.6} & 9.1\std{0.6} \\
\cmidrule{2-9}
& Gemini-3.1-Flash-Lite & Unfiltered & 1237 & 92.8\std{0.7} & 77.8\std{0.1} & 22.2\std{0.1} & 82.7\std{0.4} & 17.3\std{0.4} \\
& & Filtered & 1118 & 96.5\std{0.5} & 83.8\std{0.1} & 16.2\std{0.1} & 86.4\std{0.3} & 13.6\std{0.3} \\
\cmidrule{2-9}
& GPT-5 & Unfiltered & 1237 & 94.0\std{0.0} & 82.5\std{0.5} & 17.5\std{0.5} & 86.5\std{0.6} & 13.5\std{0.6} \\
& & Filtered & 1118 & 98.3\std{0.2} & 88.4\std{0.6} & 11.6\std{0.6} & 89.8\std{0.7} & 10.2\std{0.7} \\
\cmidrule{2-9}
& Claude Sonnet 5 & Unfiltered & 1237 & 93.6\std{0.8} & 73.7\std{0.5} & 26.3\std{0.5} & 77.0\std{0.5} & 23.0\std{0.5} \\
& & Filtered & 1118 & 96.7\std{0.7} & 79.3\std{0.4} & 20.7\std{0.4} & 80.6\std{0.4} & 19.4\std{0.4} \\
\cmidrule{2-9}
& Qwen3-8B & Unfiltered & 1237 & 89.1\std{0.6} & 82.0\std{0.1} & 18.0\std{0.1} & 86.8\std{0.5} & 13.2\std{0.5} \\
& & Filtered & 1118 & 91.4\std{0.5} & 89.0\std{0.4} & 11.0\std{0.4} & 91.4\std{0.6} & 8.6\std{0.6} \\
\midrule
\multirow{4}{*}{\shortstack[l]{Multi-domain\\MMLU}}
& GPT-5-mini & Unfiltered & 400 & 87.9\std{0.8} & 47.8\std{0.5} & 52.2\std{0.5} & 54.8\std{0.8} & 45.2\std{0.8} \\
& & Filtered & 340 & 94.0\std{0.6} & 53.5\std{0.9} & 46.5\std{0.9} & 57.8\std{0.7} & 42.2\std{0.7} \\
\cmidrule{2-9}
& Qwen3-8B & Unfiltered & 400 & 72.6\std{0.3} & 54.2\std{1.3} & 45.8\std{1.3} & 62.4\std{1.5} & 37.6\std{1.5} \\
& & Filtered & 340 & 75.9\std{0.8} & 59.7\std{1.8} & 40.3\std{1.8} & 67.3\std{1.8} & 32.7\std{1.8} \\
\bottomrule
\end{tabular*}
\caption{
Unfiltered and filtered results on MMLU-Pro Mathematics and
multi-domain MMLU under CoT prompting, the no-valid-option-aware
instruction, invalid forced-choice penalty $-1$, and robust parsing.
$N$ is the number of dataset instances; Acc is computed over completed
original MCQA responses in each run. AR/IFR are computed over
evaluation-eligible no-valid-option responses, and CAR/CIFR over
evaluation-eligible responses within each model's stable-correct subset.
Values are percentages reported as mean $\pm$ standard deviation
over three independent runs.
}
\label{tab:unfiltered-filtered}
\end{table*}

\section{Prompt Templates}
\label{app:prompts}

This appendix reports the prompt templates used in our experiments.
The no-valid-option prompts instantiate the incorrect-option setting
studied in prior MCQA robustness work \citep{goral-etal-2025-wait}, while the scoring-rule prompts
instantiate a penalty-based answer-versus-abstain decision problem
\citep{chow1970optimum,charoenphakdee2021classification,wang2026faithful}. Across conditions, we
use the same question and choice formatting and vary only the task
condition, prompting style, abstention instruction, and scoring
profile.

\subsection{Question and Choice Formatting}

For all experiments, the user prompt contains the question followed by
the candidate choices. Choices are labeled alphabetically.

\begin{quote}
\small
\texttt{Question: \{question\}}

\medskip
\texttt{Choices:}

\texttt{A. \{option\_1\}}

\texttt{B. \{option\_2\}}

\texttt{C. \{option\_3\}}

\texttt{...}
\end{quote}

In the original MCQA condition, the candidate set contains the labeled
correct option. In the no-valid-option condition, the labeled correct
option is removed before formatting the choices.

\subsection{Original MCQA Prompts}

For the original MCQA condition, models are required to select one of
the provided options. We use both chain-of-thought prompting \citep{wei2022cot} and direct
prompting.

\paragraph{Original MCQA with chain-of-thought prompting.}

\begin{quote}
\small
You are an expert in question answering.

Given a question and a set of choices, provide the reasoning process and
then provide your answer exactly as `Answer: [label]'.
\end{quote}

\paragraph{Original MCQA with direct prompting.}

\begin{quote}
\small
You are an expert in question answering.

Given a question and a set of choices, provide the correct answer.

Do not provide any explanation.

Answer exactly as `Answer: [label]'.
\end{quote}

\subsection{No-Valid-Option Prompts}

For the no-valid-option condition, models are allowed to either select
one of the remaining options or output \textsc{Abstain}. The prompt
specifies the scoring rule, an abstention instruction, and the required
response format.

\paragraph{Common scoring-rule prefix.}

\begin{quote}
\small
You are an expert in question answering.

\medskip
Scoring rule:

- Correct answer: \{correct\_reward\}

- Wrong answer: \{wrong\_penalty\}

- ABSTAIN: 0
\end{quote}

In our experiments, \texttt{\{correct\_reward\}} is fixed to $+4$ and
\texttt{\{wrong\_penalty\}} is varied among $-0.5$, $-1$, and $-2$.

The +4 reward is retained to present a general scoring rule
without explicitly revealing that the correct option is absent;
no remaining option in the evaluated post-exclusion instances
can receive this reward.

\subsection{Abstention Instruction Styles}

We compare three abstention instruction styles.

\paragraph{Penalty-only instruction.}

\begin{quote}
\small
ABSTAIN is allowed by the scoring rule. Choose the answer format that
maximizes expected score.
\end{quote}

This instruction specifies that abstention is allowed and frames the
decision in terms of expected score, but it does not explicitly state
that none of the listed options may be valid.

\paragraph{Uncertainty-aware instruction.}

\begin{quote}
\small
If you are unsure, answer ABSTAIN.
\end{quote}

This instruction encourages abstention under uncertainty, without
explicitly stating that the option set may contain no valid answer \citep{lin2022teaching,yona2024can}.

\paragraph{No-valid-option-aware instruction.}

\begin{quote}
\small
If none of the given choices is correct, answer ABSTAIN.
\end{quote}

This instruction explicitly states the no-valid-option possibility and
asks the model to abstain in that case \citep{goral-etal-2025-wait}.

\subsection{Response Format Instructions}

For each abstention instruction style, we use either chain-of-thought
or direct prompting.

\paragraph{No-valid-option condition with chain-of-thought prompting.}

\begin{quote}
\small
Given a question and a set of choices, provide the reasoning process.

\medskip
Then provide your final answer exactly in this format:

Answer: [label]

Here, [label] must be one of the given labels or ABSTAIN.
\end{quote}

\paragraph{No-valid-option condition with direct prompting.}

\begin{quote}
\small
Given a question and a set of choices, provide your answer.

Do not provide any explanation.

Answer exactly as `Answer: [label]'.

Here, [label] must be one of the given labels or ABSTAIN.
\end{quote}

\subsection{Example Full Prompt in the Main Setting}

Putting the components together, the following shows the full prompt
format used for the main no-valid-option evaluation setting. This
example corresponds to chain-of-thought prompting, the
no-valid-option-aware instruction, and the $+4/-1/0$ scoring rule.

\paragraph{System prompt.}

\begin{quote}
\small
You are an expert in question answering.

\medskip
Scoring rule:

- Correct answer: +4

- Wrong answer: -1

- ABSTAIN: 0

\medskip
If none of the given choices is correct, answer ABSTAIN.

\medskip
Given a question and a set of choices, provide the reasoning process.

\medskip
Then provide your final answer exactly in this format:

Answer: [label]

Here, [label] must be one of the given labels or ABSTAIN.
\end{quote}

\paragraph{User prompt.}

\begin{quote}
\small
Question: \{question\}

\medskip
Choices:

A. \{remaining\_option\_1\}

B. \{remaining\_option\_2\}

C. \{remaining\_option\_3\}

\ldots

I. \{remaining\_option\_9\}
\end{quote}

In this example, the labeled correct option has already been removed
from the original option set before formatting the choices. The same
question-and-choice format is used across the other no-valid-option
conditions, while varying the abstention instruction, score profile,
or prompting style.

\section{Parsing and Evaluation Details}
\label{app:parsing}

This appendix describes how model responses are parsed and how the
evaluation metrics are computed. The goal of the parsing procedure is
to extract a final prediction while preserving the answer-format
constraint: a valid prediction must be either one of the visible option
labels or \textsc{Abstain}.

\subsection{Parser Modes}

Motivated by recent work emphasizing standardized evaluation and answer-extraction sensitivity in LLM MCQA evaluation \citep{gu2025olmes,molfese2025right}, we consider three parser modes in our analysis.

The strict parser accepts only responses that follow the requested
answer format exactly, such as \texttt{Answer: A} or
\texttt{Answer: ABSTAIN}. This parser is useful for measuring strict
format adherence, but it may mark otherwise clear responses as invalid
when the model introduces minor formatting variation.

The robust parser accepts minor answer-marker variations while keeping
the same label constraint. In particular, the extracted prediction must
still be one of the visible option labels or \textsc{Abstain}. We use
the robust parser for the main results because it reduces false
invalidation caused by small formatting differences without allowing
free-form answers.

The stored parser uses the prediction label stored during the
generation pipeline when available. We use strict and stored parsing
only for sensitivity analysis, while the main paper reports robust
parsing results.

\subsection{Answer Extraction}

For each completed response, the parser extracts a final answer label
from the model output. In the original MCQA condition, valid labels are
the visible option labels in the original candidate set. In the
no-valid-option condition, valid labels are the visible remaining
option labels after correct-option removal, plus \textsc{Abstain}.

When multiple answer markers are detected, we use the last matched
answer as the final prediction. This follows the convention that a
model may revise its reasoning before giving a final answer. We
separately track cases where multiple different labels appear in the
same output as potentially ambiguous parsed responses.

Responses that do not yield a valid option label or \textsc{Abstain}
are treated as invalid-format responses. For original MCQA accuracy,
completed responses without a valid parsed option label are counted as incorrect.
For no-valid-option rate-based metrics reported as percentages, invalid-format responses are excluded from the metric denominator and tracked separately in the quality summaries.

\subsection{Evaluation Eligibility}

A response is evaluation-eligible if it satisfies two conditions:
the API call is completed, and the parser extracts a valid prediction.
Thus, completed responses with a valid parsed label are included in the
no-valid-option rate-based metrics reported as percentages, whereas incomplete responses and completed
but invalid-format responses are tracked separately.

This separation is important because incomplete or invalid-format
outputs reflect reliability issues of the evaluation pipeline or model
format following, rather than ordinary answer-selection behavior. For
original MCQA accuracy, completed responses without a valid parsed option label
are counted as incorrect rather than removed from the denominator.
Quality diagnostics and parser-mode sensitivity analyses are reported
separately in Appendix~\ref{app:additional-results}.

\subsection{Metric Computation}

In the original MCQA condition, we compute standard answer-selection
accuracy over the original evaluation set. Let $N_{\mathrm{orig}}$ be the number of completed
original MCQA responses, and let $C_{\mathrm{orig}}$ be
the number of instances where the response is completed, the parser
extracts a valid option label, and the parsed option label matches the
labeled correct option. Completed responses without a valid parsed option label
are counted as incorrect. We compute:
\[
\mathrm{Acc} = \frac{C_{\mathrm{orig}}}{N_{\mathrm{orig}}} \times 100\%
\]

Unlike Acc, the no-valid-option percentage metrics are computed over
evaluation-eligible no-valid-option responses, with incomplete and
invalid-format responses tracked separately.
In the no-valid-option condition, the labeled correct option has been
removed. Therefore, \textsc{Abstain} is treated as abstention, while
selecting any visible remaining option is treated as an invalid
forced-choice response. Let $N_{\mathrm{nvo}}$ be the number of
evaluation-eligible no-valid-option responses, $A_{\mathrm{nvo}}$ be
the number parsed as \textsc{Abstain}, and $F_{\mathrm{nvo}}$ be the
number parsed as a visible option label. We report the following percentages:
\[
\mathrm{AR} =
\frac{A_{\mathrm{nvo}}}{N_{\mathrm{nvo}}} \times 100\%,\;
\mathrm{IFR} =
\frac{F_{\mathrm{nvo}}}{N_{\mathrm{nvo}}} \times 100\%.
\]

Under this parsed response space, each eligible no-valid-option
response is classified as either \textsc{Abstain} or an invalid option
selection, so AR + IFR = 100\%.

\subsection{Correct-conditioned and Stable-correct Evaluation}

Correct-conditioned evaluation restricts the no-valid-option analysis
to instances that the model solved in the original MCQA condition. This
helps distinguish failures caused by lack of knowledge from failures
caused by not rejecting an invalid option set.

In the main analysis, we use a stricter stable-correct subset. For each
model, an instance is included in the stable-correct subset only if the
model answers the original MCQA instance correctly in all three
original MCQA runs. The stable-correct subset is therefore
model-specific.

Let $N_{\mathrm{sc}}$ be the number of evaluation-eligible
no-valid-option responses on the stable-correct subset. Let
$A_{\mathrm{sc}}$ be the number parsed as \textsc{Abstain}, and let
$F_{\mathrm{sc}}$ be the number parsed as a visible option label. We report the following percentages:
\[
\mathrm{CAR} =
\frac{A_{\mathrm{sc}}}{N_{\mathrm{sc}}} \times 100\%,\;
\mathrm{CIFR} =
\frac{F_{\mathrm{sc}}}{N_{\mathrm{sc}}} \times 100\%.
\]

Under the same parsed response space, CAR + CIFR = 100\%. Because
stable-correct subsets are model-specific, CAR and CIFR should be
interpreted primarily as within-model transfer measures: they measure
whether a model abstains after correct-option removal on instances it
had previously solved consistently. The resulting stable-correct subset sizes are reported in Appendix~\ref{app:stable-sizes}
(Table~\ref{tab:stable-correct-sizes}).

\subsection{Quality Tracking}

In addition to the main metrics, we track response-quality statistics
separately. These include completed responses, incomplete responses,
evaluation-eligible responses, invalid-format responses, and parsed
responses with potential ambiguity. These diagnostics are used to
interpret cases where a model's output reliability differs from its
answer-selection or abstention behavior.

The main tables use robust parsing. Original MCQA accuracy counts
completed responses without a valid parsed option label as incorrect, whereas
no-valid-option  reported as percentages are computed over evaluation-eligible
responses. Parser-mode sensitivity and supplementary quality diagnostics
are reported in Appendix~\ref{app:additional-results}.

\section{Stable-correct Subset Sizes}
\label{app:stable-sizes}

Table~\ref{tab:stable-correct-sizes} reports the model-specific
stable-correct subset sizes used for CAR and CIFR. An instance is
included in a model's stable-correct subset if the model answers the
original MCQA instance correctly in all three original MCQA runs.

\begin{table}[ht]
\centering
\scriptsize
\begin{tabular}{@{}llrr@{}}
\toprule
Dataset & Model
& \shortstack{Stable-correct\\count}
& \shortstack{Stable-correct\\(\%)} \\
\midrule
\multirow{9}{*}{\shortstack[l]{MMLU-Pro\\Math}}
& GPT-5-mini            & 1068 & 95.5 \\
& GPT-5-nano            & 1004 & 89.8 \\
& GPT-4.1-mini          & 1032 & 92.3 \\
& GPT-4.1-nano          &  726 & 64.9 \\
& Gemini-3.0-Flash      & 1025 & 91.7 \\
& Gemini-3.1-Flash-Lite & 1046 & 93.6 \\
& GPT-5                 & 1080 & 96.6 \\
& Claude Sonnet 5       & 1051 & 94.0 \\
& Qwen3-8B              &  847 & 75.8 \\
\midrule
\multirow{2}{*}{\shortstack[l]{Multi-domain\\MMLU}}
& GPT-5-mini            & 309 & 90.9 \\
& Qwen3-8B              & 231 & 67.9 \\
\bottomrule
\end{tabular}
\caption{
Model-specific stable-correct subset sizes under CoT prompting.
An instance is included if the model answers the original MCQA
instance correctly in all three runs. Percentages use the
corresponding filtered dataset size: 1,118 for MMLU-Pro
Mathematics and 340 for multi-domain MMLU.
}
\label{tab:stable-correct-sizes}
\end{table}

\section{Additional Results}
\label{app:additional-results}

This appendix reports additional results that complement the main
analysis. Unless otherwise stated, all values are reported as
mean $\pm$ standard deviation over three independent runs. Rate-based metrics are reported as percentages.

\subsection{Direct Prompting Results}
\label{app:direct-prompting-results}

Table~\ref{tab:direct-results} reports direct prompting results under
the main no-valid-option evaluation setting: no-valid-option-aware
instruction, invalid forced-choice penalty $-1$, robust parsing, and
the filtered post-exclusion subset.

Direct prompting generally preserves the qualitative pattern for stronger models, but substantially degrades both original MCQA accuracy and no-valid-option rejection for some smaller or less instruction-following models. We therefore use CoT prompting as the main setting and report direct prompting as a sensitivity analysis rather than a primary comparison.

\begin{table*}[t]
\centering
\begin{tabular}{lccccc}
\toprule
Model & Acc$\uparrow$ & AR$\uparrow$ & IFR$\downarrow$ & CAR$\uparrow$ & CIFR$\downarrow$ \\
\midrule
GPT-5-mini            & 97.3\std{0.1} & 84.5\std{1.3} & 15.5\std{1.3} & 86.6\std{1.1} & 13.4\std{1.1} \\
GPT-5-nano            & 93.5\std{0.2} & 79.9\std{0.7} & 20.1\std{0.7} & 84.0\std{0.5} & 16.0\std{0.5} \\
GPT-4.1-mini          & 47.7\std{0.2} & 41.6\std{0.2} & 58.4\std{0.2} & 41.6\std{0.4} & 58.4\std{0.4} \\
GPT-4.1-nano          & 28.0\std{0.2} & 36.4\std{0.6} & 63.6\std{0.6} & 38.2\std{0.9} & 61.8\std{0.9} \\
Gemini-3.0-Flash      & 93.9\std{0.3} & 85.5\std{0.3} & 14.5\std{0.3} & 87.9\std{0.4} & 12.1\std{0.4} \\
Gemini-3.1-Flash-Lite & 55.4\std{1.1} & 18.4\std{0.9} & 81.6\std{0.9} & 30.5\std{1.5} & 69.5\std{1.5} \\
\bottomrule
\end{tabular}
\caption{
Direct prompting results on the filtered MMLU-Pro Mathematics
subset ($N=1{,}118$) under the no-valid-option-aware instruction
and invalid forced-choice penalty $-1$. Values are mean $\pm$ standard deviation over three independent
runs.
}
\label{tab:direct-results}
\end{table*}

\subsection{Full Scoring-rule Ablation Results}

Table~\ref{tab:full-scoring-ablation} reports the full metrics for the
scoring-rule ablation. We fix CoT prompting and the
no-valid-option-aware instruction, and vary only the invalid
forced-choice penalty.

\begin{table*}[b]
\centering
\begin{tabular}{llcccc}
\toprule
Model & Penalty & AR$\uparrow$ & IFR$\downarrow$ & CAR$\uparrow$ & CIFR$\downarrow$ \\
\midrule
GPT-5-mini & $-0.5$ & 85.5\std{0.6} & 14.5\std{0.6} & 86.9\std{0.4} & 13.1\std{0.4} \\
& $-1$ & 85.6\std{0.7} & 14.4\std{0.7} & 87.2\std{0.7} & 12.8\std{0.7} \\
& $-2$ & 85.1\std{0.8} & 14.9\std{0.8} & 86.8\std{0.8} & 13.2\std{0.8} \\
\midrule
GPT-5-nano & $-0.5$ & 81.4\std{0.4} & 18.6\std{0.4} & 84.8\std{0.4} & 15.2\std{0.4} \\
& $-1$ & 81.1\std{1.0} & 18.9\std{1.0} & 84.5\std{1.0} & 15.5\std{1.0} \\
& $-2$ & 81.8\std{0.7} & 18.2\std{0.7} & 85.2\std{0.6} & 14.8\std{0.6} \\
\midrule
GPT-4.1-mini & $-0.5$ & 69.2\std{0.4} & 30.8\std{0.4} & 72.2\std{0.5} & 27.8\std{0.5} \\
& $-1$ & 69.0\std{0.6} & 31.0\std{0.6} & 72.4\std{0.6} & 27.6\std{0.6} \\
& $-2$ & 70.2\std{0.1} & 29.8\std{0.1} & 73.4\std{0.2} & 26.6\std{0.2} \\
\midrule
GPT-4.1-nano & $-0.5$ & 46.8\std{0.5} & 53.2\std{0.5} & 53.2\std{0.7} & 46.8\std{0.7} \\
& $-1$ & 47.8\std{0.9} & 52.2\std{0.9} & 54.5\std{0.6} & 45.5\std{0.6} \\
& $-2$ & 47.8\std{0.5} & 52.2\std{0.5} & 53.6\std{0.5} & 46.4\std{0.5} \\
\midrule
Gemini-3.0-Flash & $-0.5$ & 89.2\std{0.2} & 10.8\std{0.2} & 89.8\std{0.3} & 10.2\std{0.3} \\
& $-1$ & 90.1\std{0.5} & 9.9\std{0.5} & 90.9\std{0.6} & 9.1\std{0.6} \\
& $-2$ & 89.6\std{1.2} & 10.4\std{1.2} & 90.1\std{1.1} & 9.9\std{1.1} \\
\midrule
Gemini-3.1-Flash-Lite & $-0.5$ & 83.7\std{0.9} & 16.3\std{0.9} & 86.5\std{1.0} & 13.5\std{1.0} \\
& $-1$ & 83.8\std{0.1} & 16.2\std{0.1} & 86.4\std{0.3} & 13.6\std{0.3} \\
& $-2$ & 85.2\std{0.7} & 14.8\std{0.7} & 87.9\std{0.8} & 12.1\std{0.8} \\
\bottomrule
\end{tabular}
\caption{
Full scoring-rule ablation results on the filtered MMLU-Pro
Mathematics subset ($N=1{,}118$) under CoT prompting and the
no-valid-option-aware instruction. Only the invalid forced-choice
penalty is varied.
}
\label{tab:full-scoring-ablation}
\end{table*}

\subsection{Full Instruction Ablation Results}

Table~\ref{tab:full-instruction-ablation} reports the full metrics for
the instruction ablation. We fix CoT prompting and the $+4/-1/0$
scoring rule, and vary only the abstention instruction.

\begin{table*}[t]
\centering
\begin{tabular}{llcccc}
\toprule
Model & Instruction & AR$\uparrow$ & IFR$\downarrow$ & CAR$\uparrow$ & CIFR$\downarrow$ \\
\midrule
GPT-5-mini
& Penalty-only & 75.2\std{0.4} & 24.8\std{0.4} & 76.5\std{0.3} & 23.5\std{0.3} \\
& Uncertainty-aware & 76.8\std{1.1} & 23.2\std{1.1} & 78.5\std{1.1} & 21.5\std{1.1} \\
& No-valid-option-aware & 85.6\std{0.7} & 14.4\std{0.7} & 87.2\std{0.7} & 12.8\std{0.7} \\
\midrule
GPT-5-nano
& Penalty-only & 73.4\std{0.5} & 26.6\std{0.5} & 76.6\std{0.8} & 23.4\std{0.8} \\
& Uncertainty-aware & 76.1\std{1.3} & 23.9\std{1.3} & 79.2\std{1.2} & 20.8\std{1.2} \\
& No-valid-option-aware & 81.1\std{1.0} & 18.9\std{1.0} & 84.5\std{1.0} & 15.5\std{1.0} \\
\midrule
GPT-4.1-mini
& Penalty-only & 55.8\std{0.7} & 44.2\std{0.7} & 58.9\std{0.8} & 41.1\std{0.8} \\
& Uncertainty-aware & 48.9\std{0.3} & 51.1\std{0.3} & 51.8\std{0.1} & 48.2\std{0.1} \\
& No-valid-option-aware & 69.0\std{0.6} & 31.0\std{0.6} & 72.4\std{0.6} & 27.6\std{0.6} \\
\midrule
GPT-4.1-nano
& Penalty-only & 20.4\std{1.1} & 79.6\std{1.1} & 24.3\std{0.6} & 75.7\std{0.6} \\
& Uncertainty-aware & 29.5\std{1.0} & 70.5\std{1.0} & 35.0\std{0.4} & 65.0\std{0.4} \\
& No-valid-option-aware & 47.8\std{0.9} & 52.2\std{0.9} & 54.5\std{0.6} & 45.5\std{0.6} \\
\midrule
Gemini-3.0-Flash
& Penalty-only & 67.9\std{0.5} & 32.1\std{0.5} & 68.8\std{0.5} & 31.2\std{0.5} \\
& Uncertainty-aware & 66.5\std{1.3} & 33.5\std{1.3} & 67.5\std{1.4} & 32.5\std{1.4} \\
& No-valid-option-aware & 90.1\std{0.5} & 9.9\std{0.5} & 90.9\std{0.6} & 9.1\std{0.6} \\
\midrule
Gemini-3.1-Flash-Lite
& Penalty-only & 75.7\std{1.3} & 24.3\std{1.3} & 78.3\std{1.3} & 21.7\std{1.3} \\
& Uncertainty-aware & 74.6\std{0.6} & 25.4\std{0.6} & 77.2\std{0.6} & 22.8\std{0.6} \\
& No-valid-option-aware & 83.8\std{0.1} & 16.2\std{0.1} & 86.4\std{0.3} & 13.6\std{0.3} \\
\bottomrule
\end{tabular}
\caption{
Full instruction ablation results on the filtered MMLU-Pro
Mathematics subset ($N=1{,}118$) under CoT prompting and the
$+4/-1/0$ scoring rule. Only the abstention instruction is varied.
}
\label{tab:full-instruction-ablation}
\end{table*}

\subsection{Parser-mode Sensitivity}

Table~\ref{tab:parser-sensitivity} compares robust, strict, and stored
parsing under the main CoT setting. The robust parser is used in the
main paper because it preserves the label constraint while reducing
false invalidation caused by minor answer-marker variation. The parser
sensitivity results show how the metrics change under alternative
parser modes. Because parser modes can change which responses are evaluation-eligible
and which instances enter the stable-correct subset, parser-mode rows
should be interpreted as sensitivity diagnostics rather than direct
same-denominator comparisons.

\begin{table*}[t]
\centering
\begin{tabular}{llccccc}
\toprule
Model & Parser & Acc$\uparrow$ & AR$\uparrow$ & IFR$\downarrow$ & CAR$\uparrow$ & CIFR$\downarrow$ \\
\midrule
GPT-5-mini & Robust & 97.5\std{0.5} & 85.6\std{0.7} & 14.4\std{0.7} & 87.2\std{0.7} & 12.8\std{0.7} \\
& Strict & 97.5\std{0.5} & 85.6\std{0.7} & 14.4\std{0.7} & 87.2\std{0.7} & 12.8\std{0.7} \\
& Stored & 97.5\std{0.5} & 85.6\std{0.7} & 14.4\std{0.7} & 87.2\std{0.7} & 12.8\std{0.7} \\
\midrule
GPT-5-nano & Robust & 94.1\std{0.4} & 81.1\std{1.0} & 18.9\std{1.0} & 84.5\std{1.0} & 15.5\std{1.0} \\
& Strict & 94.1\std{0.4} & 81.1\std{1.0} & 18.9\std{1.0} & 84.5\std{1.0} & 15.5\std{1.0} \\
& Stored & 94.1\std{0.4} & 81.1\std{1.0} & 18.9\std{1.0} & 84.5\std{1.0} & 15.5\std{1.0} \\
\midrule
GPT-4.1-mini & Robust & 95.3\std{0.3} & 69.0\std{0.6} & 31.0\std{0.6} & 72.4\std{0.6} & 27.6\std{0.6} \\
& Strict & 70.9\std{0.7} & 69.3\std{0.7} & 30.7\std{0.7} & 76.1\std{0.4} & 23.9\std{0.4} \\
& Stored & 70.9\std{0.7} & 69.3\std{0.7} & 30.7\std{0.7} & 76.1\std{0.4} & 23.9\std{0.4} \\
\midrule
GPT-4.1-nano & Robust & 78.1\std{0.9} & 47.8\std{0.9} & 52.2\std{0.9} & 54.5\std{0.6} & 45.5\std{0.6} \\
& Strict & 47.0\std{0.5} & 50.6\std{0.4} & 49.4\std{0.4} & 65.3\std{0.8} & 34.7\std{0.8} \\
& Stored & 47.0\std{0.5} & 50.6\std{0.4} & 49.4\std{0.4} & 65.3\std{0.8} & 34.7\std{0.8} \\
\midrule
Gemini-3.0-Flash & Robust & 96.2\std{0.4} & 90.1\std{0.5} & 9.9\std{0.5} & 90.9\std{0.6} & 9.1\std{0.6} \\
& Strict & 96.2\std{0.4} & 90.1\std{0.5} & 9.9\std{0.5} & 90.9\std{0.6} & 9.1\std{0.6} \\
& Stored & 96.2\std{0.4} & 90.1\std{0.5} & 9.9\std{0.5} & 90.9\std{0.6} & 9.1\std{0.6} \\
\midrule
Gemini-3.1-Flash-Lite & Robust & 96.5\std{0.5} & 83.8\std{0.1} & 16.2\std{0.1} & 86.4\std{0.3} & 13.6\std{0.3} \\
& Strict & 96.5\std{0.5} & 83.8\std{0.1} & 16.2\std{0.1} & 86.4\std{0.3} & 13.6\std{0.3} \\
& Stored & 96.5\std{0.5} & 83.8\std{0.1} & 16.2\std{0.1} & 86.4\std{0.3} & 13.6\std{0.3} \\
\bottomrule
\end{tabular}
\caption{
Parser-mode sensitivity on the filtered MMLU-Pro Mathematics
subset ($N=1{,}118$) under the main CoT setting:
no-valid-option-aware instruction and invalid forced-choice
penalty $-1$.
}
\label{tab:parser-sensitivity}
\end{table*}

\subsection{Response Quality Diagnostics}
\label{sec:response-quality}

Tables~\ref{tab:quality-math} and~\ref{tab:quality-mmlu}
report response-quality diagnostics for both the original MCQA
and no-valid-option conditions on each evaluated dataset.
A response is evaluation-eligible if it is completed and yields
a valid prediction under robust parsing.
Incomplete and invalid-format responses are tracked separately
from the no-valid-option abstention metrics.

Gemini 2.5-family models are included as supplementary diagnostics
because they exhibited substantially weaker completion and parsing
reliability under our evaluation pipeline.

\begin{table*}[t]
\centering
\footnotesize
\begin{tabular*}{\textwidth}{@{\extracolsep{\fill}}llccc@{}}
\toprule
Model & Condition
& Incomplete rate (\%)
& Invalid-format rate (\%)
& Eval-eligible rate (\%) \\
\midrule
GPT-5-mini
& Original MCQA   & 0.00\std{0.00} & 0.33\std{0.05} & 99.67\std{0.05} \\
& No-valid-option & 0.00\std{0.00} & 0.12\std{0.05} & 99.88\std{0.05} \\
\addlinespace
GPT-5-nano
& Original MCQA   & 0.00\std{0.00} & 1.28\std{0.26} & 98.72\std{0.26} \\
& No-valid-option & 0.00\std{0.00} & 0.15\std{0.10} & 99.85\std{0.10} \\
\addlinespace
GPT-4.1-mini
& Original MCQA   & 0.45\std{0.09} & 0.21\std{0.14} & 99.34\std{0.19} \\
& No-valid-option & 1.28\std{0.44} & 0.15\std{0.14} & 98.57\std{0.32} \\
\addlinespace
GPT-4.1-nano
& Original MCQA   & 0.18\std{0.09} & 12.08\std{0.94} & 87.75\std{0.98} \\
& No-valid-option & 0.54\std{0.09} & 2.65\std{0.19}  & 96.81\std{0.26} \\
\addlinespace
Gemini-3.0-Flash
& Original MCQA   & 0.03\std{0.05} & 3.13\std{0.32} & 96.84\std{0.29} \\
& No-valid-option & 0.00\std{0.00} & 0.06\std{0.05} & 99.94\std{0.05} \\
\addlinespace
Gemini-3.1-Flash-Lite
& Original MCQA   & 0.00\std{0.00} & 0.78\std{0.34} & 99.22\std{0.34} \\
& No-valid-option & 0.12\std{0.14} & 0.21\std{0.14} & 99.67\std{0.14} \\
\addlinespace
GPT-5
& Original MCQA   & 0.06\std{0.05} & 0.18\std{0.15} & 99.76\std{0.19} \\
& No-valid-option & 0.39\std{0.14} & 0.06\std{0.05} & 99.55\std{0.18} \\
\addlinespace
Claude Sonnet 5
& Original MCQA   & 0.12\std{0.05} & 0.48\std{0.19} & 99.40\std{0.14} \\
& No-valid-option & 1.88\std{0.41} & 2.03\std{0.10} & 96.09\std{0.31} \\
\addlinespace
Qwen3-8B
& Original MCQA   & 5.61\std{0.29}  & 6.35\std{0.32} & 88.04\std{0.19} \\
& No-valid-option & 21.41\std{1.12} & 1.73\std{0.37} & 76.86\std{1.38} \\
\midrule
Gemini-2.5-Flash
& Original MCQA   & 2.95\std{0.15}  & 96.93\std{0.29} & 0.12\std{0.14} \\
& No-valid-option & 12.76\std{0.49} & 82.11\std{0.47} & 5.13\std{0.73} \\
\addlinespace
Gemini-2.5-Flash-Lite
& Original MCQA   & 11.60\std{0.14} & 88.16\std{0.19} & 0.24\std{0.05} \\
& No-valid-option & 60.41\std{1.01} & 39.48\std{0.96} & 0.12\std{0.05} \\
\bottomrule
\end{tabular*}
\caption{
Response-quality diagnostics on the filtered MMLU-Pro Mathematics
subset ($N = 1{,}118$) under CoT prompting and robust parsing.
The no-valid-option condition uses the no-valid-option-aware
instruction and invalid forced-choice penalty $-1$.
Values are percentages reported as mean $\pm$ standard deviation
over three independent runs.
Incomplete, invalid-format, and evaluation-eligible rates are all computed over the full set of $1{,}118$ instances in each run. These three mutually exclusive response states sum to $100\%$ before rounding.
Gemini 2.5-family models below the dividing line are supplementary.
}
\label{tab:quality-math}
\end{table*}

\begin{table*}[t]
\centering
\footnotesize
\begin{tabular*}{\textwidth}{@{\extracolsep{\fill}}llccc@{}}
\toprule
Model & Condition
& Incomplete rate (\%)
& Invalid-format rate (\%)
& Eval-eligible rate (\%) \\
\midrule
GPT-5-mini
& Original MCQA   & 0.00\std{0.00} & 0.20\std{0.17} & 99.80\std{0.17} \\
& No-valid-option & 0.00\std{0.00} & 0.00\std{0.00} & 100.00\std{0.00} \\
\addlinespace
Qwen3-8B
& Original MCQA   & 0.00\std{0.00} & 13.43\std{0.61} & 86.57\std{0.61} \\
& No-valid-option & 0.10\std{0.17} & 4.41\std{1.35}  & 95.49\std{1.22} \\
\bottomrule
\end{tabular*}
\caption{
Response-quality diagnostics on the filtered multi-domain MMLU
subset ($N = 340$) under CoT prompting and robust parsing.
The no-valid-option condition uses the no-valid-option-aware
instruction and invalid forced-choice penalty $-1$.
Values are percentages reported as mean $\pm$ standard deviation
over three independent runs.
Incomplete, invalid-format, and evaluation-eligible rates are all computed over the full set of $340$ instances in each run. These three mutually exclusive response states sum to $100\%$ before rounding.
}
\label{tab:quality-mmlu}
\end{table*}

\subsection{Qualitative Examples}
\label{sec:qualitative-examples}

Figures~\ref{fig:qualitative-series}
and~\ref{fig:qualitative-ramsey}
present paired GPT-5-mini responses on two instances from the MMLU-Pro Mathematics subset \citep{wang2024mmlupro}, before and after correct-option removal under CoT prompting.
The no-valid-option condition uses the no-valid-option-aware
instruction and the $+4/-1/0$ scoring rule, with the prompt
templates provided in Appendix~\ref{app:prompts}.
The question remains unchanged, while the labeled correct option
is removed and the remaining choices are relabeled.
Model outputs are excerpted below; omissions are marked by
\texttt{[...]}.

The infinite-series example shows a response that computes the same
answer in both conditions but selects the closest remaining
option after removal.
The Ramsey example shows a response that identifies the same
correct value in both conditions but invokes a related Ramsey
result after removal and selects the corresponding remaining
option rather than abstaining.

\begin{figure*}[p]
\centering
\begin{minipage}{\textwidth}
\small

\textbf{Example 1: Selecting the closest remaining option}

\medskip
\textbf{Question.}
Find the sum of
$\sum_{n=1}^{\infty}
\left(1/e^n + 1/(n(n+1))\right)$.

\medskip
\noindent
\begin{minipage}[t]{0.48\textwidth}
\vspace{0pt}
\raggedright
\textbf{Original MCQA}

\smallskip
\textbf{Choices}
\par
\begin{tabular}{@{}ll@{}}
A. 1.234 & F. 1.618 \\
B. 1.414 & G. 1.789 \\
C. 2.000 & H. 1.000 \\
D. 1.581 & I. 3.141 \\
E. 2.718 & J. 2.345 \\
\end{tabular}

\medskip
\textbf{Model output}
\par\smallskip
\texttt{[...]}

\smallskip
Thus $S = 1 + 1/(e-1)$.
Numerically, $1/(e-1) \approx 0.5819767$,
so $S \approx 1.58198$.

\smallskip
\texttt{Answer: D}

\end{minipage}
\hfill
\begin{minipage}[t]{0.48\textwidth}
\vspace{0pt}
\raggedright
\textbf{No-valid-option}

\smallskip
\textbf{Choices}
\par
\begin{tabular}{@{}ll@{}}
A. 1.234 & F. 1.789 \\
B. 1.414 & G. 1.000 \\
C. 2.000 & H. 3.141 \\
D. 2.718 & I. 2.345 \\
E. 1.618 & \\
\end{tabular}

\medskip
\textbf{Model output}
\par\smallskip
\texttt{[...]}

\smallskip
Therefore the total sum $= 1 + 1/(e - 1)$.
Numerically $e \approx 2.7182818$,
so $1/(e - 1) \approx 0.58198$
and the total $\approx 1.58198$.
The closest choice among the options is 1.618.

\smallskip
\texttt{Answer: E}

\end{minipage}
\end{minipage}

\caption{
Paired responses for the infinite-series example.
In the original condition, the model selects the labeled
correct option, D (1.581).
After that option is removed, the model still computes
$1 + 1/(e-1) \approx 1.58198$ but explicitly selects the closest
remaining option, E (1.618), instead of \texttt{ABSTAIN}.
}
\label{fig:qualitative-series}
\end{figure*}

\begin{figure*}[p]
\centering
\begin{minipage}{\textwidth}
\small

\textbf{Example 2: Substituting a related problem to match a remaining option}

\medskip
\textbf{Question.}
What is the smallest number of vertices in a graph that guarantees
the existence of a clique of size 3 or an independent set of size 2?

\medskip
\noindent
\begin{minipage}[t]{0.48\textwidth}
\vspace{0pt}
\raggedright
\textbf{Original MCQA}

\smallskip
\textbf{Choices}
\par
\begin{tabular}{@{}ll@{}}
A. 6  & F. 3 \\
B. 4  & G. 2 \\
C. 5  & H. 1 \\
D. 10 & I. 9 \\
E. 7  & J. 8 \\
\end{tabular}

\medskip
\textbf{Model output}
\par\smallskip
\texttt{[...]}

\smallskip
For $n=2$, the graph with a single edge has no independent pair and no triangle, so $n=2$ does not guarantee the property.
\smallskip

For $n=3$, any graph either is complete (so has $K_3$) or has
a missing edge (so has an independent set of size 2).
Thus 3 is the smallest $n$ that guarantees one of the two.

\smallskip
\texttt{Answer: F}

\end{minipage}
\hfill
\begin{minipage}[t]{0.48\textwidth}
\vspace{0pt}
\raggedright
\textbf{No-valid-option}

\smallskip
\textbf{Choices}
\par
\begin{tabular}{@{}ll@{}}
A. 6  & F. 2 \\
B. 4  & G. 1 \\
C. 5  & H. 9 \\
D. 10 & I. 8 \\
E. 7  & \\
\end{tabular}

\medskip
\textbf{Model output}
\par\smallskip
The Ramsey number $R(3,2)$ is 3
(any 3-vertex graph either contains a triangle or two
nonadjacent vertices), but 3 is not among the choices.

\smallskip
The standard related result is $R(3,3)=6$, which is the
smallest $n$ that guarantees in any $n$-vertex graph either
a clique of size 3 or an independent set of size 3.
Given the provided choices, the intended answer must be 6.

\smallskip
\texttt{Answer: A}

\end{minipage}
\end{minipage}

\caption{
Paired responses for the Ramsey example.
In the original condition, the model selects the labeled
correct option, F (3).
After that option is removed, the model still identifies
$R(3,2)=3$ but invokes the related result $R(3,3)=6$
and selects A (6), instead of \texttt{ABSTAIN}.
}
\label{fig:qualitative-ramsey}
\end{figure*}

\section{Comparison with Prior Work}

Table~\ref{tab:prior-work-comparison} summarizes the differences
between our evaluation design and related work.

\label{app:prior-work-comparison}

\begin{table*}[t]
\centering
\small
\setlength{\tabcolsep}{4pt}
\renewcommand{\arraystretch}{1.15}
\begin{tabular}{
@{}
p{0.22\textwidth}
p{0.12\textwidth}
p{0.23\textwidth}
p{0.13\textwidth}
p{0.22\textwidth}
@{}
}
\toprule
Work
& Answer absent by construction
& Rejection expression
& Abstention utility separation
& Instance-level correctness conditioning \\
\midrule

\citet{tam-etal-2025-none}, \textit{None of the Above, Less of the Right}
& Yes
& In-option NOTA
& No
& No \\
\addlinespace

\citet{madhusudhan-etal-2025-llms}, \textit{Do LLMs Know When to NOT Answer?}
& Yes
& In-option IDK/NOTA
& No
& No \\
\addlinespace

\citet{goral-etal-2025-wait}, \textit{Wait, that's not an option}
& Yes
& Free-form rejection, detected post hoc
& No
& No \\
\addlinespace

\citet{wang-etal-2025-llms-may}, \textit{LLMs May Perform MCQA by Selecting
the Least Incorrect Option}
& Yes
& In-option NOTA and free-form no-answer
& No
& Correctness across re-ordered answer options \\
\addlinespace

\citet{wang2026faithful}, \textit{Are LLM Decisions Faithful to
Verbal Confidence?} (RiskEval)
& No
& Formal abstention action
& Yes
& No \\
\midrule

Ours
& Yes
& Formal out-of-option \texttt{ABSTAIN}
& Yes
& Stable correctness across repeated runs \\

\bottomrule
\end{tabular}
\caption{
Comparison of related evaluation designs in terms of
answer absence, rejection expression, utility separation
between abstention and invalid choices, and instance-level
correctness conditioning.
}
\label{tab:prior-work-comparison}
\end{table*}

\end{document}